\documentclass{elsarticle}
\journal{Elsevier}

\begin{document}
% Commenting the line below out will disable hyperref ... sometimes this is a problem when building pdf
%\include{00README.XXX}
\begin{frontmatter}

% Journal paper title

\title{Vehicle classification using ResNets, localisation and spatially-weighted pooling}

\author{Rohan Watkins}
\author{Nick Pears} 
\author{Suresh Manandhar}
\address{Department of Computer Science, University of York, York, UK}
\begin{abstract}
We investigate whether ResNet architectures can outperform more traditional Convolutional Neural Networks on the task of fine-grained vehicle classification. We train and test ResNet-18, ResNet-34 and ResNet-50 on the Comprehensive Cars dataset without pre-training on other datasets. We then modify the networks to use Spatially Weighted Pooling. Finally, we add a localisation step before the classification process, using a network based on ResNet-50. We find that using Spatially Weighted Pooling and localisation both improve classification accuracy of ResNet50. Spatially Weighted Pooling increases accuracy by 1.5 percent points and localisation increases accuracy by 3.4 percent points. Using both increases accuracy by 3.7 percent points giving a top-1 accuracy of 96.351\% on the Comprehensive Cars dataset. Our method achieves higher accuracy than a range of methods including those that use traditional CNNs. However, our method does not perform quite as well as pre-trained networks that use Spatially Weighted Pooling.
\end{abstract}


\end{frontmatter}

\section{Introduction}

\label{sec: intro}
In the fine-grained vehicle classification problem, a class consists of both make and model attributes, with the optional addition of the year that a particular model version was released (e.g. Ford Focus 2003, 2008). If such a `year' attribute is required, the difficulty of the problem increases significantly, due to the similarity of updated models.
This problem differs from more coarse recognition, which may categorise by vehicle type (car, van, bus, etc) and have far fewer classes. Several methods have been used to try and solve fine-grained vehicle classification. The main limitation of these approaches is the inability to differentiate between similar car models. Recently, \cite{He2016DeepRL} proposed residual networks (ResNets). They have shown state-of-the-art performance in general image classification, as well as other computer vision tasks. In this paper, we investigate their performance in fine-grained vehicle classification, using the publicly-available \emph{Comprehensive Cars} dataset, also investigating the use of a localisation network and spatially weighted pooling.

\section{Related literature on vehicle classification}

\cite{DBLP:conf/cvpr/YangLLT15} use the CNN OverFeat model presented by \cite{Overfeat2014}, which has 5 convolutional layers, max pooling layers after the 1st, 2nd and 5th convolutional layers and 3 fully connected layers. The CNN has 145 million parameters and is pretrained on the ImageNet classification task. To adapt it to vehicle classification, it is fine-tuned on their \emph{Comprehensive Cars} dataset. This consists of 214,345 images of 1687 car models from multiple perspectives. To train and test the network they used a subset of the data consisting of 431 car models
Here, the provided training dataset contains 16,016 images and the test dataset has 14,939 images from these classes. 
Classification was done as make and model, ignoring the year the model was released. They achieved top-1 accuracy of 76.7\% and top 5 accuracy of 91.7\%. In the latest revision of their paper they also report results for AlexNet, \cite{Alexnet2012}, and GoogLeNet, \cite{GoogleNet2015}. AlexNet achieved a top-1 of 81.9\% and a top-5 of 94.0\%. GoogLeNet achieved a top-1 of 91.2\% and a top-5 of 98.1\%
Their results, along with their dataset, provide a good baseline for future experiments by the research community in fine-grained vehicle classification. 

\cite{Wang2016Finegrained} use a parts-based method to classify vehicles by make, model and year. They note that often the only difference between models of different years is from very specific parts such as lights. To address this issue their parts-based method combines global and local features in conjunction with a novel voting mechanism. In the first step, a trained \emph{Strongly Supervised Deformable Parts Model} (SSDPM), by \cite{azizpour2012}, is used to detect and localise the seven defined car parts. 
Then a histogram of gradients feature is extracted to represent the global appearance of the image. A CNN is used to extract local features from each car part. The CNN used is a version of AlexNet, which has 5 convolutional layers and 3 fully connected layers. In total the CNN has about 60 million parameters and the features are extracted after the second fully connected layer. 
An SVM is then used to produce a probability vector using the local and global features. Final fine-grained classification is then done with a novel voting mechanism that uses the probability vector produced by the SVM and a set of weights, learned by SVM, that describe the importance of the preliminary recognition results. 
The dataset used consists of 4584 images of 50 types of vehicle from 8 manufacturers. The images are occlusion-free car fronts, labelled with manufacturer, model and year. They achieved results of 94.2\% on annotated data and 92.3\% on SSDPM localised data. This outperformed the standard Caffe method which achieved 87.1\% and 88.9\% respectively.

\cite{Chabot2017DeepEI} employ 3D vehicle models and edge-colour invariant features. They use a dataset of 3D vehicle models to generate images for training a neural network. Using a 3D dataset allows a lot of images to be generated from any desired angle. This is useful for vehicle classification as the vehicle dataset contain images from multiple viewpoints. Another advantage of generating images from 3D models is that there aren’t any labelling errors, which are more likely when dealing with fine-grained classification. GoogLeNet is trained on edge map images and colour images to find a set of features that are common to both types of image. 
These learnt features are referred to as edge-colour invariant features. Both types of image are used for training because they found that training on only the edge maps from the generated images led to overfitting on edge structures that only appeared in the generated images and not in real images.
For training they used 36,451 images from Comprehensive Cars and 124,416 images generated from 3D models. GoogLeNet was pretrained on ImageNet and then fine-tuned on the vehicle datasets. They trained networks on colour and edge images separately and together. 
They test their networks on the Comprehensive Cars test set, 15,626 images over 431 classes. The networks trained on only colour or only edge images performed poorly on the type of data they weren’t trained on. The network trained on colour images performed very poorly on edge images (3.8\%) but the network trained on edge images performed better on colour images (46.7\%). This suggests that features learned on edge images somewhat generalise to colour images, but features learnt on colour images don’t generalise to edge images. On the colour test set the network trained on colour and edge datasets achieves top-1 accuracy of 95.9\% and top-5 of 99.4\%. 

\cite{Dehghan2017ViewIV} use a localisation step before fine-grained classification has this has produced excellent performance. \cite{Hu2017LocationAware} use two methods, multi-task learning and multi-scale pre-processing as well as localisation before classification. Two neural networks are used, one for localisation and one for classification, both of which are based on AlexNet. By determining the position of the vehicle, a tight crop can be used to limit the impact of noisy backgrounds on classification. Also, if the size of the vehicle in the image is small (i.e. it is far away) detecting its location means it can be scaled to a larger size. To train the localisation network, additional data with saliency maps are used as the vehicle datasets don’t have saliency maps. Saliency maps are like bounding boxes but are not restricted to be a rectangle. The networks are trained using a multi-task strategy so that the networks are trying to optimise on two or more problems at the same time. The localisation network is trained using two loss functions. A loss function for the produced saliency map and one for the produced bounding box. The network for classification also has two loss functions, one for the error in class prediction and one for view point prediction. Before the classification network is trained all training, images are input into the localisation network. The images are then cropped based on the output of the localisation network and resized to 400 by 400 pixels. The images also have scale and shift pre-processing applied before being input into the classification network. To test the effects of multi-task learning and multi-scale pre-processing, they use the Comprehensive Cars dataset and train models with and without pretraining on ImageNet. They use two training splits of the Comprehensive Cars dataset, a 50:50 split and a 70:30 split. The 50:50 split has 16,016 training images and 14,939 test images of 431 classes. The 70:30 split has 36,456 training images and 15,627 test images. They also perform tests with and without scale pre-processing and multi-tasking. Their results showed that using either proposed method improves performance and using both together improves performance further. They also showed that using the localisation network improved accuracy compared to only using AlexNet. Using both methods, multi-tasking and multi-scale, without pretraining and training on the 50:50 split the network achieved a top-1 accuracy of 77.19\% and top-5 of 91.44\%. With pre-training, fine-tuning on the 50:50 split, and using both methods the top-1 accuracy improved to 91\% and the top-5 to 97.77\%. Using pretraining, the 70:30 split, and both methods gave the best results achieving a top-1 of 94.3\% and top-5 of 98.9\%.

\subsection{Discussion}

All the papers discussed use networks pretrained on the ImageNet dataset. The networks are then fine-tuned on the Comprehensive Cars dataset apart from in \cite{Wang2016Finegrained}, where they use a custom data set. The results reported by the papers are also on the Comprehensive Cars dataset, again excluding \cite{Wang2016Finegrained}, which make them directly comparable.
The networks used in the papers discussed above are OverFeat, AlexNet and GoogLeNet. OverFeat and AlexNet have 8 layers while GoogLeNet has 22. OverFeat has the most parameters, 145 million, AlexNet has 61 million and GoogLeNet has 6 million. When using the same approach (same dataset, training, testing) GoogLeNet achieves the highest accuracy (91.2\%), \cite{DBLP:conf/cvpr/YangLLT15}. This suggests that network depth plays a key role in accuracy, which is supported by \cite{He2016DeepRL}, and that having more parameters doesn’t necessarily improve the accuracy of the network.

Both \cite{Chabot2017DeepEI}  and \cite{Hu2017LocationAware} take a multi-faceted approach to classification. In \cite{Chabot2017DeepEI}, they train using both colour and edge images and in \cite{Hu2017LocationAware} they use a network for localisation trained on salient maps and bounding boxes. Both these methods outperform the standard approach used in \cite{DBLP:conf/cvpr/YangLLT15}. In \cite{Chabot2017DeepEI}, GoogLeNet is used to learn edge-colour invariant features, using this method the accuracy improved from 91.2\% to 95.9\%. This is an interesting result as the convolutional layers should be capable of learning filters to extract features related to edges without the images being fist converted into edge maps. This suggests that when training on only colour images the network finds that at least some of the edge features learnt from training on edge maps aren’t as discriminative as other features present in the colour images. It is possible that if a larger network was used, and so had more learnable filters, it would learn to extract the edge features. 
In \cite{Hu2017LocationAware} AlexNet is used with multi-task learning, multi-scale image pre-processing and localisation. Using this method, the accuracy improved from 81.9\% to 94.3\%. This is only 1.6 points lower than the edge-colour method which used GoogLeNet. The results suggest that this method offers a better improvement in accuracy (12.4 points vs 4.7 points) but as different networks were used, it isn’t a direct comparison. In summary, using deeper networks, multi-task learning, multi-scale image pre-processing and localisation can increase the accuracy of vehicle classification compared to the original benchmark results on the \emph{Comprehensive Cars} dataset of \cite{DBLP:conf/cvpr/YangLLT15}.

\section{Network architecture}

The residual networks that we used are based on the Keras implementation by \cite{KerasResNet}. This implementation provides several residual networks including ResNet-18, ResNet-34 and ResNet-50. The implementation uses the improved residual blocks as described by \cite{He2016DeepRL}. ResNet-18 and 34 use the standard architecture but ResNet-50 uses the bottleneck architecture. 
%A summary of the networks is given in Table 6 and 
We now describe ResNet-34 in more detail to illustrate how the networks are structured. ResNet-34 is a 34-layer network with 21 million parameters, it uses shortcut connections every two layers. The input to the network is a 224x224 image with 3 colour channels (RGB). The output is a 431-dimensional vector that is a one-hot encoding of the classes. The class labels are converted into one-hot encoded vectors using the utility functions provided by Keras. In finer detail, it has an initial convolutional layer with a filter size of 7x7 and 64 filters. This is followed by a max pooling layer with a pool size of 3x3 and a stride of 2. It then has 4 blocks of convolutional layers with filters of size 3x3. The first block has 6 layers each with 64 filters. The second block has 8 layers each with 128 filters. The third block has 12 layers each with 256 filters. The final block has 6 layers each with 512 filters. After the final block there is an average pooling layer with a pool size of 7x7 and a step of 1. The final layer is the classifier layer which is a fully connected layer with 431 neurons (one per class). The figure taken from , shows the structure of ResNet-34, the activation and batch normalisation layers are excluded for simplicity but are used before each convolutional layer, as in \cite{He2016DeepRL}.

The networks used for localisation have the same structure as those used for classification, described above, except they have four classification layers. The four classification layers are connected to the average pooling layer which provides a copy of its output to each classification layer. Of the four classification outputs, the two for predicting the centre point have 25 nodes and the two for predicting width and height have 40 nodes. Each node represents a bin of 7 pixels, so the network can predict location from 0 to 175, and sizes from 0 to 280. The predicted value is taken as the value in the middle of the bin. We modified our networks, as suggested by \cite{Hu2017LocationAware}, when adding a spatially weighted pooling layer. Firstly, the average pooling layer and fully connected layer are removed. Secondly, the SWP layer is added followed by a batch normalisation layer, a 1024 node fully connected layer and then the final classification layer. We implemented the SWP layer in Keras by creating a subclass of the Layer class which is the superclass of all layers used by Keras. 

\begin{figure}[!t]
\centering
\includegraphics[width=0.9\textwidth]{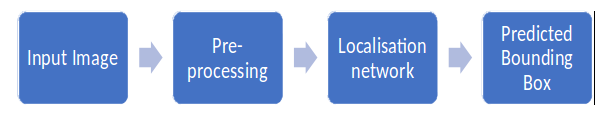} \\
\includegraphics[width=0.9\textwidth]{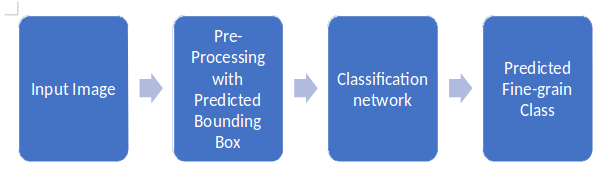}
\caption{Two-stage approach: localisation then classification}
\label{fig:twostage}
\end{figure}

Figure \ref{fig:twostage} gives an overview of how the localisation and classification networks are used in the implementation. First, the localisation network receives a pre-processed image as input and outputs a predicted bounding box. Then the unprocessed input image is pre-processed with the predicted bounding box and input into the classification network, which predicts a fine-grained class.

\section{Training}

In this work, we use images from the Comprehensive Cars dataset to construct a new training split that contains 84,623 images from 841 classes. The new training set has roughly 100 images per class and does not contain any of the images from the Comprehensive Cars test set. We use this dataset to initially train the networks instead of using a network pre-trained on ImageNet as was done in \cite{DBLP:conf/cvpr/YangLLT15}, \cite{Hu2017LocationAware}, and \cite{Sochor2017}. We use transfer learning with the original Comprehensive Cars training set of 16,016 images (431 classes) so that the networks can be tested against the test set of 14,939 images (and the same 431 classes).

\section{Evaluation}

\subsection{Evaluation process and metrics}

When designing the evaluation process, the key concern was ensuring that our results were comparable to existing results on the Comprehensive Cars test dataset. The results reported in  \cite{DBLP:conf/cvpr/YangLLT15} are Top-1 and Top-5 percentage accuracy for model classification. As such, we report the same statistical performance metrics. For testing networks without localisation, a single central crop and a single scale (256) was used instead of using the 10-crop method used to test residual networks in \cite{He2016DeepRL}. For testing networks with localisation, the predicted bounding boxes are used to crop the images first, then they are scaled so the largest side is 224 pixels. The width and height of the bounding boxes are increased by 10\% before cropping. This is done to reduce the amount of vehicle lost in the case that the predicted bounding box is too small. We have aimed to keep the testing process as similar to others in the literature using the Comprehensive Cars dataset, namely \cite{DBLP:conf/cvpr/YangLLT15}, \cite{Chabot2017DeepEI}, \cite{Hu2017LocationAware}, \cite{HuWLS17}, in order that any difference in reported results are due to the architecture, pre-processing and training of the networks. 

\subsection{Results}

Table \ref{table:localisation} contains the classification accuracies of the localisation networks on the Comprehensive Cars dataset. The accuracy is classification accuracy because the localisation networks use bins of values instead of regression. The networks have four classification outputs, centre x, centre y, width, height, so the accuracy of each individual output is reported. The x and y outputs have 25 bins of 7 pixels. The width and height outputs have 40 bins of 7 pixels. The results show that ResNet50 achieves the best accuracy and adding the Spatially-Weighted Pooling (SWP) layer reduces the mean accuracy by almost 1 percentage point.

\begin{table*}[!t]
\caption{Classification accuracy (\%) of our localisation networks. The percentage accuracies are top-1 for each of Centre X, Centre Y, Width, Height. The final column gives the mean percentage accuracy of the four outputs.}
\centering
\begin{tabular}{|c|c|c|c|c|c|}
\hline
Method & Cent. X & Cent. Y  & Wid.  & Hght.  & Mean  \\ \hline \hline
ResNet50 & 85.354 & 87.380 & 77.723 & 81.095 & 82.888 \\ \hline
ResNet50+SWP & 84.999 & 86.226 & 75.872 & 80.499 & 81.889 \\ \hline
\end{tabular}
\label{table:localisation}
\end{table*}

Table \ref{table:classification} contains the top-1 and top-5 classification accuracies achieved using several methods on the Comprehensive Cars dataset. The best accuracy was achieved using ResNet50 with both SWP and localisation. The results show that using localisation and SWP individually improve the accuracy achieved by ResNet50. Additionally, using both together further increases the accuracy. Our method achieves higher accuracies than all methods discussed in the literature, that published results on the Comprehensive Cars dataset, except the pretrained ResNet50 with SWP and ResNet101 with SWP which achieved  top-1 accuracies of 97.5\% and 97.6\% respectively, \cite{HuWLS17}. 
The edge-colour method proposed by \cite{Chabot2017DeepEI} is the highest accuracy method uses a traditional CNN. Our best method achieves an accuracy 0.451 percent points higher, and correctly classifies an additional 67 images of the 14939-image test dataset. 

\begin{table*}[!t]
\caption{Classification accuracy of our localisation networks. The accuracies are top-1 for each of Centre X, Centre Y, Width, Height. The final column gives the mean accuracy of the four outputs.}
\centering
\begin{tabular}{|c|c|c|}
\hline
Method & Top-1 (\%) & Top-5 (\%) \\ \hline \hline
\multicolumn{3}{|c|}{Ours}  \\\hline
ResNet18 & 78.132 & 95.108  \\ \hline
ResNet34 & 84.516 & 97.171  \\ \hline
ResNet50 & 92.622 & 98.907  \\ \hline
ResNet50 with SWP & 94.125 & 99.074 \\\hline
ResNet50 with Loc & 96.050 & 99.423 \\\hline
ResNet50 with SWP and Loc & 96.351  & 99.463\\\hline
\multicolumn{3}{|c|}{Other literature}  \\\hline
AlexNet & 81.9 & 94.0  \\\hline
GoogleNet & 91.2 & 98.1  \\\hline
OverFeat & 76.7 & 91.7  \\\hline
Edge-Colour & 95.9 & 99.4  \\\hline
Location-Aware, Multi-Task & 94.3 & 98.9  \\\hline
Pre-trained ResNet50 with SWP & 97.5 & -  \\\hline
Pre-trained ResNet101 with SWP & 97.6  & -  \\\hline

\end{tabular}
\label{table:classification}
\end{table*}

Using Spatially Weighted Pooling increases the accuracy of ResNet50 by 1.503 percent points from 92.662\% to 94.125\%. Using \emph{localisation} increases the accuracy of ResNet50 by 3.428 percent points to 96.050\%; using localisation gives a greater increase in accuracy than SWP. When localisation is used with ResNet50 with SWP, the accuracy only increases by 0.301 percent points from 96.050\% to 96.351\%. The small difference between ResNet50 with localisation and ResNet50 with SWP and localisation suggests that localisation and SWP are providing a similar function. The slight increase suggests that SWP does provide some extra function that localisation doesn’t. The addition of SWP increases the top-1 accuracy more than the top-5 accuracy suggesting that it allows the network to better distinguish between certain similar models but doesn’t help to classify a small percent of images where the actual class isn’t predicted in the top-5. As the accuracy increases you are left with the most difficult images to classify, neither localisation or SWP provide the functionality needed to classify the remaining 3.649\% of the test set (545 images).

\subsection{Processing speed}

The processing speed of our method is reported in Table \ref{table:speed} as the number of images processed per second (frames per second -FPS). The FPS is reported using a batch size of 1 and 32 on a desktop PC with a Nvidia 8GB GTX 1060 and is calculated by predicting classes of 10,000 images. A batch size of 32 is also included because a typical surveillance system would have multiple cameras and so would be processing multiple images at a time. Our method is capable of processing 83 FRS without localisation and 42 FPS with localisation. Our results for the basic ResNet50  are lower than those reported using a GTX 1080 as expected because the GTX 1060 is less powerful, \cite{Sochor2017}. Table \ref{table:speed} shows that using the SWP layer has negligible effect on the processing speed, while using localisation has a large effect, as expected, because two networks are being used instead of one. Depending on the vehicle surveillance application, and the available hardware, the large decrease in processing speed may outweigh the increase in accuracy.

\begin{table*}[!t]
\caption{Processing speed of our method with a batch size of 1 and 32. Speed reported as frames per second (FPS).}
\centering
\begin{tabular}{|c|c|c|}
\hline
Method & FPS (Batch Size 1) &  FPS (Batch Size 32) \\ \hline \hline
ResNet50 & 44 & 83  \\ \hline
ResNet50 with SWP & 44 & 83 \\\hline
ResNet50 with SWP and Loc & 21  & 42 \\\hline
\end{tabular}
\label{table:speed}
\end{table*}

\section{Discussion}

A common issue faced by fine-grained vehicle classification systems, including ours, is the classification of highly similar car models. Table \ref{tab:carimages} shows example images of the three most commonly misclassified cars with the top row being the input and the bottom row being the predicted class. In all these cases the classification method can predict the correct make, but a highly similar model is predicted instead of the correct model. The results show that the top-5 accuracies are above 99\%. Thus the correct model is almost always predicted when extending to the 5 most probable models, but it is difficult for the networks to distinguish between very similar models. Our method often confuses the BMW X3 and BMW X5, which are very similar and as the example images in Table \ref{tab:carimages} show it is not easy to distinguish between them.

\begin{table*}[!t]
\caption{Model classification errors for (left-to-right) BMW X5, Mitsubishi Lancer and mercedes Benz S-class. Top row : input images; bottom row, an example image consistent with the erroneous predicted class label.}
\label{tab:carimages}
\centering
\begin{tabular}{|c|c|c|}
\hline
\includegraphics[width=0.3\textwidth]{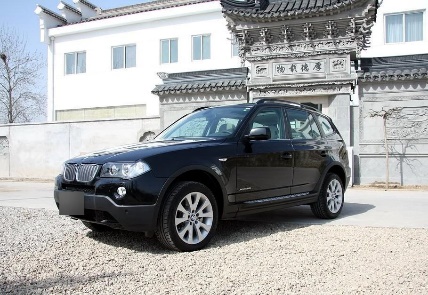} & \includegraphics[width=0.3\textwidth]{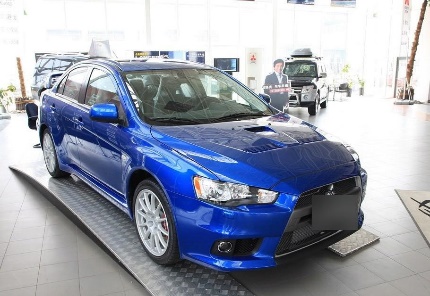}   & \includegraphics[width=0.3\textwidth]{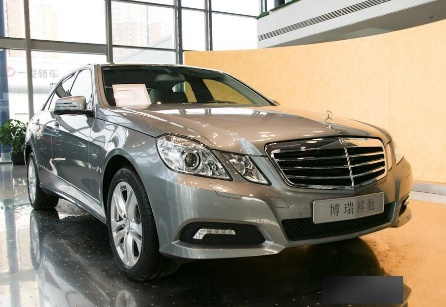}  \\ \hline 
\includegraphics[width=0.3\textwidth]{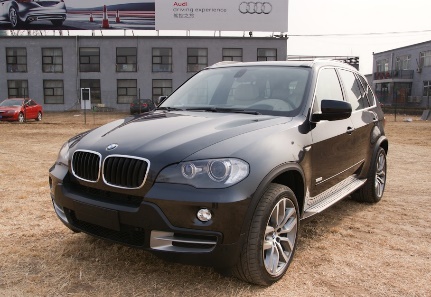} & \includegraphics[width=0.3\textwidth]{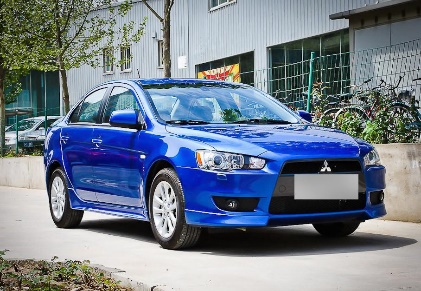}   & \includegraphics[width=0.3\textwidth]{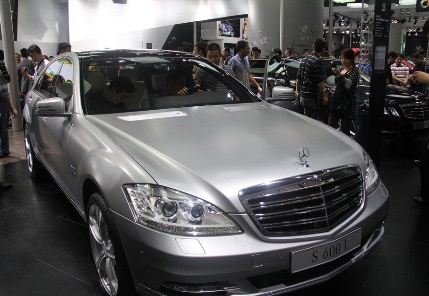}  \\ \hline 
\end{tabular}
\end{table*}

\subsection{Limited input image resolution}

Our method uses a single image of size 224x224 as input, this has several drawbacks when trying to classify vehicles that are very similar to other vehicles. Highly similar models have few distinguishing features and even then, those can be very slight. It is possible that given a single image it might be impossible to tell two models apart because the distinguishing features either aren’t present in the image or are too similar at the relatively low resolution of the input. For example, a distinguishing feature might be a small difference in the shape of a headlight which is not visible at lower image resolutions.  However, the advantage of using a single small image as input is that the network is relatively small and computationally inexpensive, thus there is an accuracy-speed trade-off.

\subsection{Input image type}

The dataset used to train and test our method consists of web-nature images and not surveillance-nature images. Datasets such as \emph{boxcars116k}, which consist of surveillance images are more challenging, the same methods achieve worse accuracy on it than on \emph{Comprehensive Cars}. Our method should perform well on surveillance data if transfer learning is used to adapt our networks or if they are retrained. This is because the localisation step should normalise the size of the vehicle in the image and reduce the amount of background which is often noisy in surveillance data. However, our localisation method can only handle images with a single vehicle in which limits generalisability.

\subsection{Pretraining}

The networks used in our implementations are not pretrained on the ImageNet dataset. An advantage of this is that the training time is much shorter. Each network converges in less than 100 iterations using the Comprehensive Cars dataset. However, a disadvantage is that the accuracy achieved without pretraining is lower. Using ResNet50 with SWP and without pretraining we achieved a top 1 accuracy of 94.125\% compared to 97.5\% achieved with pre-training, as reported by Qichang [40]. This difference of 3.375\% equates to 504 images of the 14939-image test set. This suggests that our best method, ResNet50 with localisation and SWP, could be improved by pretraining the networks.
% potentially surpassing 97.5\% accuracy.

\subsection{Spatially-weighted pooling}

As discussed, we added SWP layers to residual networks as they have been shown to improve accuracy. We use a fully connected layer with 1024 nodes and a SWP layer with 9 masks which was shown to offer the best accuracy in \cite{HuWLS17}. 
We found that this does improve the accuracy of ResNet50 when it is used for classification. However, when we modified our ResNet50 localisation network in the same way it reduced the accuracy. This could be because the number of nodes in the fully connected layer and the number of masks used is not suitable for the number of classes. The classification network has 431 classes while the localisation network has 130. The localisation network also has multiple outputs which could be a contributing factor.

To visualise what the SWP layer is producing, a heatmap of a selection of its outputs for specific inputs is shown in Table \ref{tab:heatmaps}. It shows the output for a BMW X3, BMW X5, Mitsubishi Evo and a Mitsubishi Lancer EX, which are all commonly misclassified. The images were formed by flattening the output of the SWP layer to create a heatmap where lighter areas correspond to higher values. The areas on the heatmaps don’t directly correspond to locations of the images as the SWP layer has 9, 7 by 7, masks that are applied to 2048 features.  Table \ref{tab:heatmaps} shows that the outputs are highly similar with all having two primary areas of activation, along the very top and about a 3rd of the way up from the bottom. Activation is largely in the same two bands which suggests that they are encoding the most common image features. The areas of low activation likely encode less common features and features that the SWP masks have learnt to suppress to improve classification accuracy. For example, the masks could be suppressing features related to the image background. It is also possible that the reason for areas of low activation is because there are too many masks in the SWP layer for the number of feature maps produced by the convolutional layers. Even if too many masks have been used using fewer masks will have negligible effect on network size and computation cost. Alternatively the networks could be pre-trained on ImageNet as in \cite{HuWLS17}, which may allow the layers preceding the SWP layer to learn more powerful features that the SWP layer can then make better use of.

\begin{table*}[!t]
\caption{Heat map of activations. Top row : BMW : X3 (left) X5 (right); bottom row : Mitsubishi : evo (left), lancer (right).  }
\label{tab:heatmaps}
\centering
\begin{tabular}{|c|c|}
\hline
\includegraphics[width=0.3\textwidth]{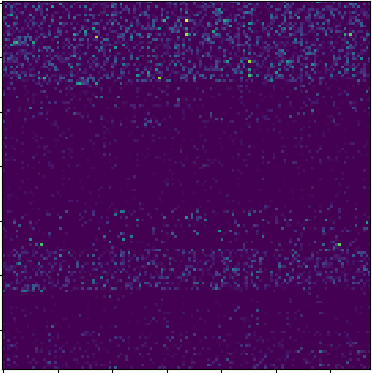} &  \includegraphics[width=0.3\textwidth]{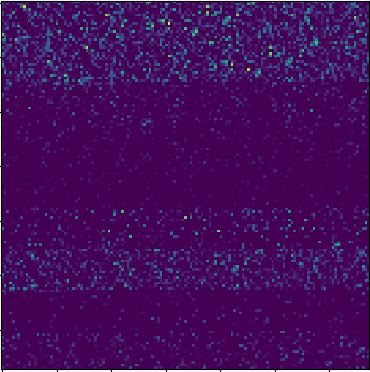}  \\ \hline 
\includegraphics[width=0.3\textwidth]{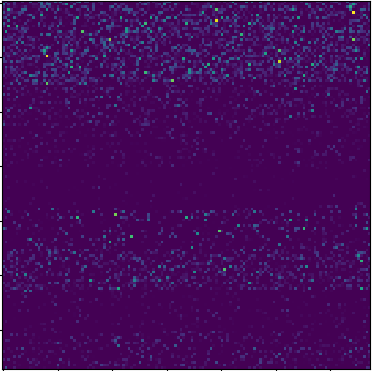}   & \includegraphics[width=0.3\textwidth]{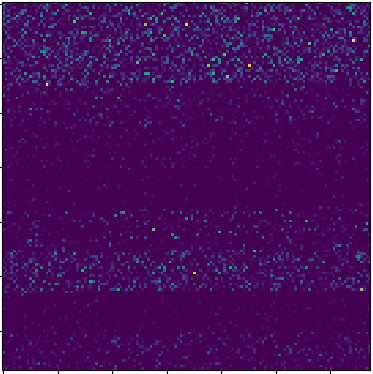}  \\ \hline 
\end{tabular}
\end{table*}

\subsection{Localisation}

Our localisation method can achieve an average accuracy of 82.9\% which is sufficient for normalising the size of the vehicles in the images and cropping them to remove background. Our method uses bins of values instead of a direct regression to make the localisation more robust at the expense of precision. Different numbers of bins are used for the location and size to better fit the data. For the width and height, 40 bins are used, and for the centre x and y position 25 bins are used.

\begin{figure}[!t]
\centering
\includegraphics[width=0.9\textwidth]{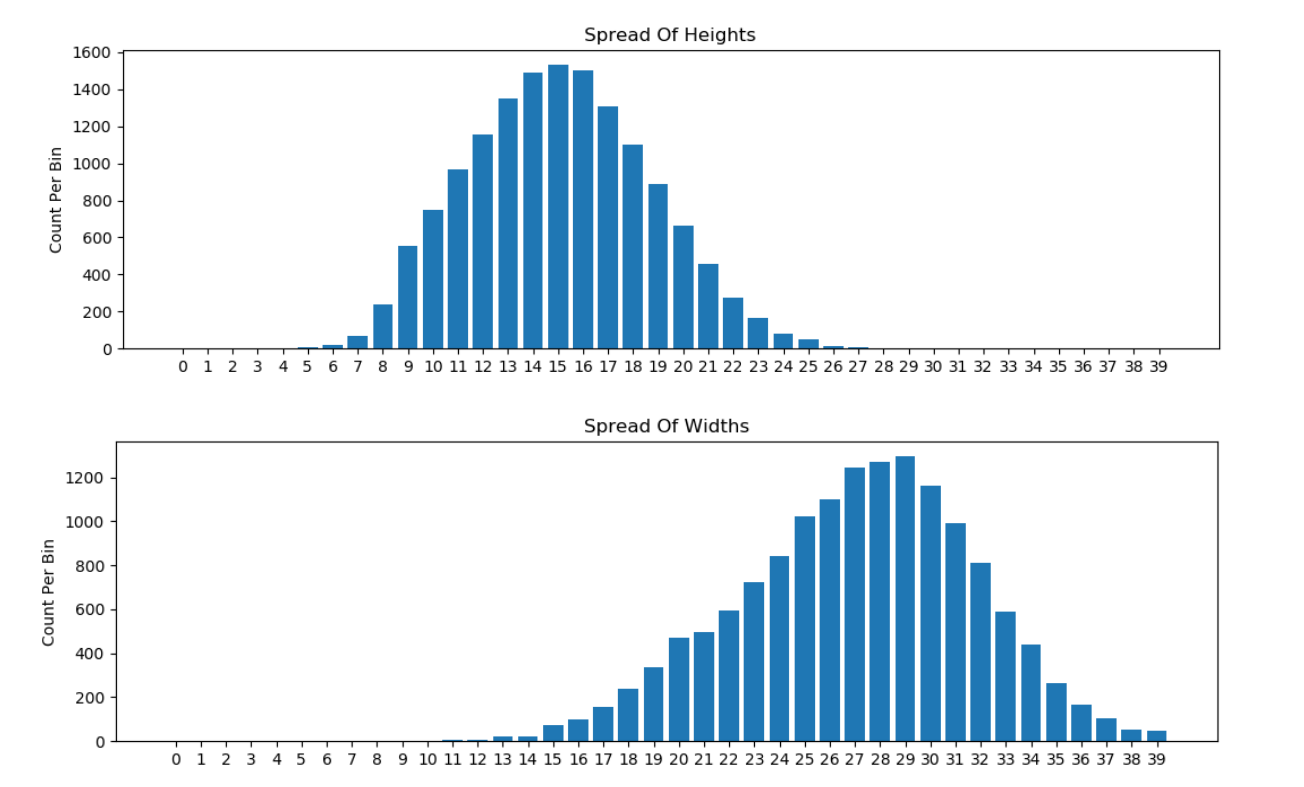}
\caption{Histograms for widths and heights of bounding boxes}
\label{fig:widthHeight}
\end{figure}

Figure \ref{fig:widthHeight} shows the distribution of the widths and heights for the bounding boxes in the training dataset with pre-processing applied. Part of the image pre-processing is to randomly rescale the images by randomly selecting a scale. This along with natural variation in the dataset causes the spread of widths and heights across the bins. For the heights the bins on the lower and upper extremes don’t have any images in and for the widths the lower bins don’t have any images in. For the width there are images with bounding box widths greater than those that fit into the last bin. This happens when an image with a large bounding box width gets resized to a large scale. When this happens, the width is added to the last bin. Both these problems could be solved by normalising the width and heights such that they are centred on the middle bin. 

\begin{figure}[!t]
\centering
\includegraphics[width=0.9\textwidth]{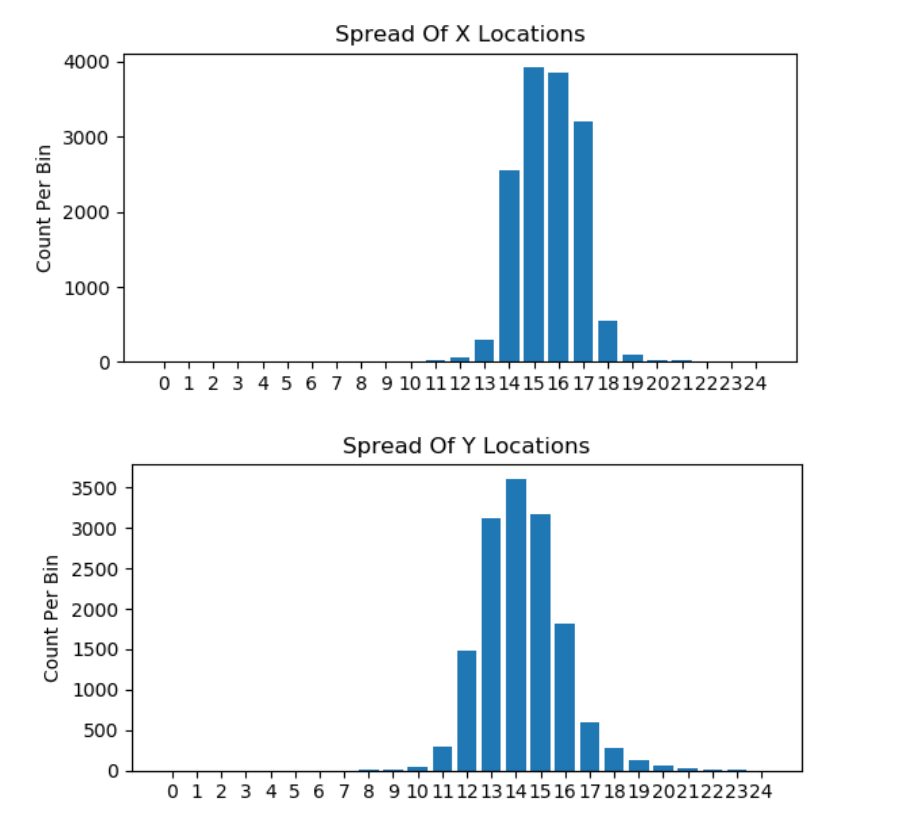}
\caption{Histograms for locations of bounding boxes}
\label{fig:loc}
\end{figure}

Figure \ref{fig:loc} shows the distribution of the centre x and y locations for the bounding boxes in the training dataset with pre-processing applied. In the pre-processing a 224x224 crop of the image is taken from a random location. Even with this pre-processing there isn’t much variation is the centre location. Over 50\% of x locations are in bin 15 and 16 with most remaining locations split between bin 14 and 17. Similarly, over 50\% of the y locations fall into bins 13, 14 and 15. Neither the x or y location spread is centred on the middle bin. This shows that location of the vehicles in the images are most likely to be centred towards the lower right. The graphs also show that our network wasn’t trained with vehicles around the edges of the input image so it likely to perform poorly in those cases. There are several unused bins resulting in redundancy within the neural network which could be removed by using fewer bins and normalising the data.

The localisation method has lower accuracy for predicting width and height than for the x and y location. The width and height have a larger spread than the x and y locations which could attribute to the lower prediction accuracy.  Another reason that width and height accuracy is lower could be that part of the vehicle is often cropped out by the pre-processing, so the bounds of the vehicle aren’t always in the image. This means that the network must predict the size based on partial information.

%Figure X shows that 
We find that when the localisation method incorrectly predicts a value it most commonly only 1 bin away. This means that the incorrect predicted values are mostly less than 1.5 $\times$ bin size pixels, 10.5 pixels, away from the true value. There are predictions that are more than 3 bins away but these account for less than 1\% of the total predictions. Our method is therefore robust, but not highly precise. Using smaller bins would allow for more precise predictions but is likely to make it less robust, as the data would be spread over more narrower bins. This could cause a decrease in accuracy like the one between the height prediction and x location prediction. A limitation of our method is that it only predicts a single bounding box. This was sufficient for the Comprehensive Cars dataset as it only contains images of single vehicles. More complex localisation, 
such as YOLO 9000, \cite{Redmon2017}, could be used to predict multiple bounding boxes for images with more than one vehicles in.

\subsection{Network size}

Our vehicle classification methods are based on ResNet50 which has 50 layers and 24 million parameters. Modifying ResNet50 to have a 1024 node fully connected layer and SWP layer increases the number of parameters to 43 million, with the majority coming from the fully connected layer and only 441 coming from the SWP layer. Although adding SWP does drastically increase the number of parameters it does also increase the accuracy of the network, so it is a trade-off between accuracy and computational cost. The localisation network is also based on ResNet50 so has 50 layers and 24 million parameters. Our best method uses the classification network with SWP and the localisation network. The best performing method, reported in the literature review, that doesn’t use residual networks is the deep edge-colour method proposed by \cite{Chabot2017DeepEI}. Their method uses a pretrained GoogLeNet, which has 11 million parameters and is 22 layers deep. Our method is more accurate by 0.451 percent points but uses two networks. This means that our solution has a total of 67 million parameters compared to 11 million. Our method is therefore more computationally expensive but more accurate. Using pretraining with our method could improve the accuracy without increasing the number of parameters in the networks.

\section{Conclusions}

We have explored fine-grained vehicle recognition using the Comprehensive Cars dataset using a set of ResNet architectures of different sizes, without using pretraining. Also, we have investigated the performance improvements to be gained with network additions for both localisation and spatially-weighted pooling. Our best system uses both of these additions with ResNet50, giving a 96.351\% and 99.463\% accuracy in top-1 and top-5 metrics respectively. Our system runs at 42 FPS using batches of 32 images when running on a desktop PC with a Nvidia 8GB GTX 1060.

\bibliographystyle{model2-names}
\bibliography{refs}

\begin{thebibliography}{10}
\expandafter\ifx\csname url\endcsname\relax
  \def\url#1{\texttt{#1}}\fi
\expandafter\ifx\csname urlprefix\endcsname\relax\def\urlprefix{URL }\fi
\expandafter\ifx\csname href\endcsname\relax
  \def\href#1#2{#2} \def\path#1{#1}\fi

\bibitem{He2016DeepRL}
K.~He, X.~Zhang, S.~Ren, J.~Sun, Deep residual learning for image recognition,
  2016 IEEE Conference on Computer Vision and Pattern Recognition (CVPR) (2016)
  770--778.

\bibitem{DBLP:conf/cvpr/YangLLT15}
L.~Yang, P.~Luo, C.~C. Loy, X.~Tang, A large-scale car dataset for fine-grained
  categorization and verification, in: {CVPR}, {IEEE} Computer Society, 2015,
  pp. 3973--3981.

\bibitem{Overfeat2014}
P.~Sermanet, D.~Eigen, X.~Zhang, M.~Mathieu, R.~Fergus, Y.~Lecun, Overfeat:
  Integrated recognition, localization and detection using convolutional
  networks, in: International Conference on Learning Representations
  (ICLR2014), CBLS, April 2014, 2014.

\bibitem{Alexnet2012}
A.~Krizhevsky, I.~Sutskever, G.~E. Hinton, Imagenet classification with deep
  convolutional neural networks, in: Proceedings of the 25th International
  Conference on Neural Information Processing Systems - Volume 1, NIPS'12,
  2012, pp. 1097--1105.

\bibitem{GoogleNet2015}
C.~Szegedy, W.~Liu, Y.~Jia, P.~Sermanet, S.~E. Reed, D.~Anguelov, D.~Erhan,
  V.~Vanhoucke, A.~Rabinovich, Going deeper with convolutions, in: {CVPR},
  {IEEE} Computer Society, 2015, pp. 1--9.

\bibitem{Wang2016Finegrained}
Q.~Wang, Z.~Wang, J.~Xiao, J.~Xiao, W.~Li, Fine-grained vehicle recognition in
  traffic surveillance, in: 17th Pacific-Rim Conference on Advances in
  Multimedia Information Processing - Volume 9916, PCM 2016, 2016, pp.
  285--295.

\bibitem{azizpour2012}
H.~Azizpour, I.~Laptev, {Object Detection Using Strongly-Supervised Deformable
  Part Models}, in: A.~Fitzgibbon, S.~Lazebnik, P.~Perona, Y.~Sato, C.~Schmid
  (Eds.), {ECCV 2012 - European Conference on Computer Vision}, LNCS - Lecture
  Notes in Computer Science, {Springer}, Florence, Italy, 2012, pp. 836--849.

\bibitem{Chabot2017DeepEI}
F.~Chabot, M.~A. Chaouch, J.~Rabarisoa, C.~Teuli{\`e}re, T.~Chateau, Deep
  edge-color invariant features for 2d/3d car fine-grained classification, 2017
  IEEE Intelligent Vehicles Symposium (IV) (2017) 733--738.

\bibitem{Dehghan2017ViewIV}
A.~Dehghan, S.~Z. Masood, G.~Shu, E.~G. Ortiz, View independent vehicle make,
  model and color recognition using convolutional neural network, CoRR
  abs/1702.01721.

\bibitem{Hu2017LocationAware}
B.~Hu, J.-H. Lai, C.-C. Guo, Location- aware fine- grained vehicle type
  recognition using multi- task deep networks, Neurocomputing 243 (2017)
  60--68.

\bibitem{KerasResNet}
Kotikalapudi, \href{https://github.com/raghakot/keras-resnet}{Keras resnet}
  (2018).
\newline\urlprefix\url{https://github.com/raghakot/keras-resnet}

\bibitem{Sochor2017}
J.~Sochor, J.~Spanhel, A.~Herout, Boxcars: Improving fine-grained recognition
  of vehicles using 3-d bounding boxes in traffic surveillance, {IEEE} Trans.
  Intelligent Transportation Systems xx~(99) (2017) 1--12.

\bibitem{HuWLS17}
Q.~Hu, H.~Wang, T.~Li, C.~Shen, Deep cnns with spatially weighted pooling for
  fine-grained car recognition, {IEEE} Trans. Intelligent Transportation
  Systems 18~(11) (2017) 3147--3156.

\bibitem{Redmon2017}
J.~Redmon, A.~Farhadi, Yolo9000: Better, faster, stronger, in: 2017 IEEE
  Conference on Computer Vision and Pattern Recognition (CVPR), 2017, pp.
  6517--6525.

\end{thebibliography}

\end{document}